\documentclass[runningheads]{llncs}
\usepackage{graphicx}
\begin{document}
\title{Model Explainability in Physiological and Healthcare-based Neural Networks}
\author{Rohit Sharma\inst{1} \and
Abhinav Gupta\inst{2} \and
Arnav Gupta\inst{2} \and
Bo Li\inst{2}}
\authorrunning{R. Sharma et al.}
\institute{University of Kentucky, USA \and
Tencent, USA\\
\email{\{boli, agupta\}@tencent.com}}
\maketitle
\begin{abstract}
The estimation and monitoring of SpO2 are crucial for assessing lung function and treating chronic pulmonary diseases. The COVID-19 pandemic has highlighted the importance of early detection of changes in SpO2, particularly in asymptomatic patients with clinical deterioration. However, conventional SpO2 measurement methods rely on contact-based sensing, presenting the risk of cross-contamination and complications in patients with impaired limb perfusion. Additionally, pulse oximeters may not be available in marginalized communities and undeveloped countries. To address these limitations and provide a more comfortable and unobtrusive way to monitor SpO2, recent studies have investigated SpO2 measurement using videos. However, measuring SpO2 using cameras in a contactless way, particularly from smartphones, is challenging due to weaker physiological signals and lower optical selectivity of smartphone camera sensors. The system includes three main steps: 1) extraction of the region of interest (ROI), which includes the palm and back of the hand, from the smartphone-captured videos; 2) spatial averaging of the ROI to produce R, G, and B time series; and 3) feeding the time series into an optophysiology-inspired CNN for SpO2 estimation. Our proposed method can provide a more efficient and accurate way to monitor SpO2 using videos captured from consumer-grade smartphones, which can be especially useful in telehealth and health screening settings.

\end{abstract}

\section{Introduction}

The estimation and monitoring of SpO2 are crucial for assessing lung function and treating chronic pulmonary diseases. The COVID-19 pandemic has highlighted the importance of early detection of changes in SpO2, particularly in asymptomatic patients with clinical deterioration. However, conventional SpO2 measurement methods rely on contact-based sensing, presenting the risk of cross-contamination and complications in patients with impaired limb perfusion. Additionally, pulse oximeters may not be available in marginalized communities and undeveloped countries.

To address these limitations and provide a more comfortable and unobtrusive way to monitor SpO2, recent studies have investigated SpO2 measurement using videos. However, measuring SpO2 using cameras in a contactless way, particularly from smartphones, is challenging due to weaker physiological signals and lower optical selectivity of smartphone camera sensors. Previous methods have relied on explicit feature extraction as a variant of pulse oximeter principles.

In this study, we propose an innovative approach to SpO2 measurement using explainable convolutional neural network (CNN) models. Our approach extracts features holistically from all three color channels of contactless videos captured using consumer-grade smartphone cameras. By using CNN models, we can address the challenges of contactless SpO2 measurement and improve the accuracy and efficiency of monitoring SpO2. This approach has the potential to be adopted in health screening and telehealth, particularly in areas where conventional contact-based methods and pulse oximeters are not widely available.

 In recent years, there has been an increasing interest in using videos to monitor SpO2 in a non-invasive and unobtrusive way that could be adopted in health screening and telehealth. However, measuring SpO2 using cameras in a contactless way, especially from smartphones, presents challenges due to weaker physiological signals and the lower optical selectivity of smartphone camera sensors. Most prior studies have resorted to explicit feature extraction methods as a variant of the principle of pulse oximeters. In contrast, this paper introduces explainable convolutional neural network (CNN) models that extract features from all three color channels holistically for SpO2 measurement from contactless videos captured using consumer-grade smartphone cameras. The proposed method aims to address the limitations of previous approaches and provide a more accurate and efficient way to monitor SpO2 using videos. 

The proposed system for contactless SpO2 monitoring using CNN from smartphone videos is presented in Figure 1. The system includes three main steps: 1) extraction of the region of interest (ROI), which includes the palm and back of the hand, from the smartphone-captured videos; 2) spatial averaging of the ROI to produce R, G, and B time series; and 3) feeding the time series into an optophysiology-inspired CNN for SpO2 estimation. The proposed CNN is designed based on the light-tissue interaction principle applied to physiological sensing, which enhances its explainability. The hand region is considered as a proof-of-concept in this work, as it raises fewer privacy concerns compared to using the face for SpO2 measurement, which is a common approach in prior studies. Additionally, recording hand videos is a safer way for health condition screening and data collection during the COVID-19 pandemic, in accordance with mask-wearing guidelines. The contributions of the proposed work are outlined later in the paper.

\begin{enumerate}
    \item We take a data-driven approach and use visualization of the weights for the RGB channel combinations to demonstrate the explainability of our model. We show that the choice of color band learned by the neural network is consistent with the suggested color bands used in optophysiological methods. 
    \item We analyze the impact of the two sides of the hand and different skin tones on the quality of SpO2 estimation. Our findings indicate that skin tone can affect the performance of the SpO2 estimation, and the back of the hand may provide more accurate results than the palm. 
    \item We compare our optophysiologically inspired neural network structures with the state-of-the-art neural network structure designed for this problem and achieve more accurate SpO2 prediction. Our proposed method can provide a more efficient and accurate way to monitor SpO2 using videos captured from consumer-grade smartphones, which can be especially useful in telehealth and health screening settings.
\end{enumerate}

\begin{figure*}
    \centering
    \includegraphics[scale=0.6]{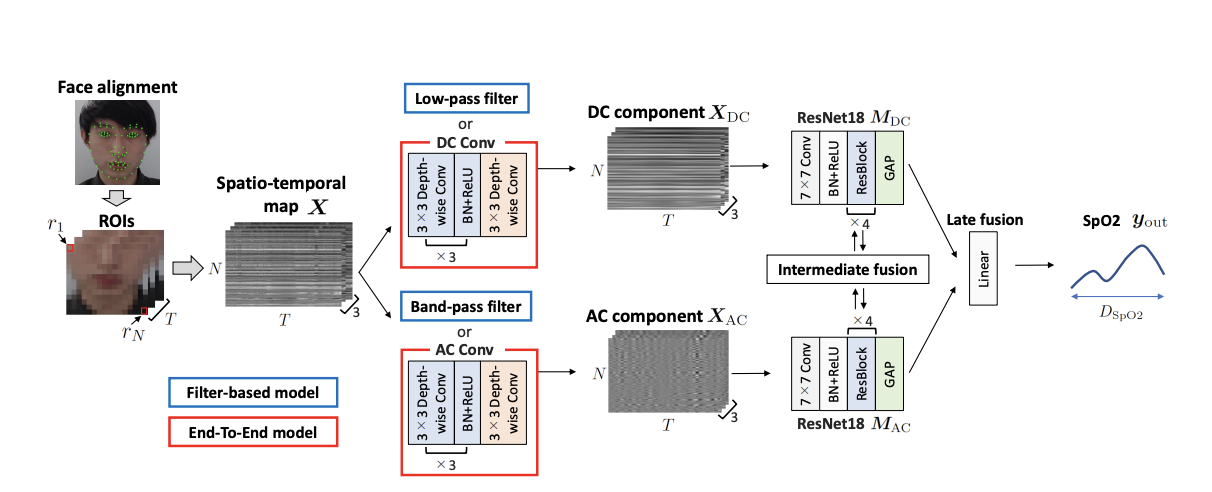}
    \caption{The proposed method involves constructing a spatio-temporal map and extracting DC and AC components using filtering processes or convolutional layers. }
    \label{fig:side2side}
\end{figure*}

\begin{figure*}
    \centering
    \includegraphics[scale=0.75]{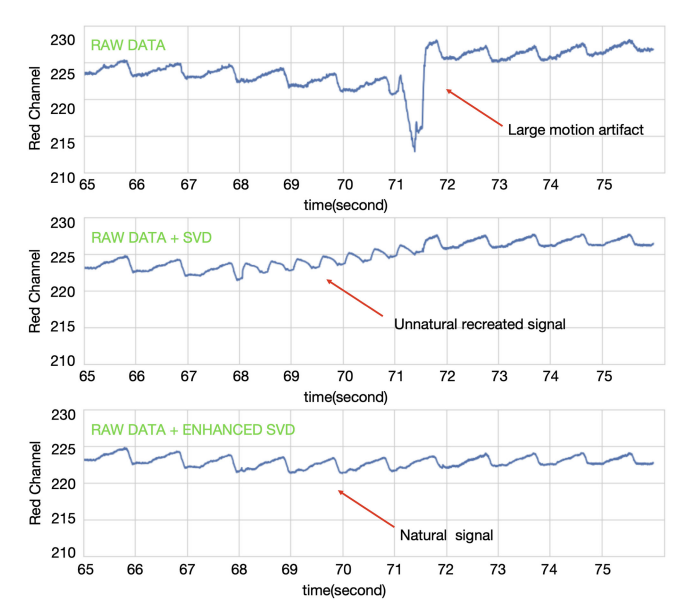}
    \caption{Signal Comparison of different artifact methods. The propsoed method is at hte bottom.}
    \label{fig:side2side}
\end{figure*}

The human body's hemoglobin (Hb) protein carries oxygen from the lungs to the body's tissues. The level of blood oxygen saturation (SpO2) is a measure of the proportion of oxygenated hemoglobin (HbO2) to total hemoglobin in the blood and indicates how well the respiratory system is functioning [1]. SpO2 falls within the normal range of 95\% to 100\% [1]. SpO2 is a critical indicator of the body's ability to meet metabolic demands, and a drop in SpO2 can signal inadequate oxygenation and clinical deterioration [1]. Pulse oximetry is a convenient and noninvasive way to measure SpO2 continuously [4].

Pulse oximeters operate on the ratio-of-ratios principle, which was first proposed by Aoyagi in the 1970s [4]. These devices emit light at two wavelengths (conventionally red and infrared) through the fingertip, and the light interacts with and is attenuated by the blood and tissue. An optical sensor receives the transmitted light, which provides information about pulsatile blood volume. The pulsatile blood volume at the two wavelengths is processed to estimate SpO2. Specifically, SpO2 is roughly proportional to the ratio-of-ratios, which is computed as the ratio between the pulsatile/AC and relatively stationary/DC components of the transmitted pulse signals at the red and infrared wavelengths [4]. Pulse oximeters are commonly used in hospitals, clinics, and homes.

The level of blood oxygen saturation (SpO2) is an important indicator of respiratory function, representing the ratio of oxygenated hemoglobin to total hemoglobin in the blood. An abnormal drop in SpO2 can indicate inadequate oxygenation and clinical deterioration, making it a critical parameter to monitor. Pulse oximeters, which utilize the ratio-of-ratios method, are commonly used to measure SpO2 noninvasively and continuously, but contact-based measurements can be uncomfortable and may raise sanitation concerns. Therefore, non-contact video-based SpO2 estimation methods have been investigated to mitigate these issues [10-30]. Over the years, DeepNeural Networks have been used extensively to remotely obtain the heart rate from facial video [5-10, 27-33].

Existing non-contact methods can be grouped into two categories based on the setup of cameras and light sources. The first category uses monochromatic sensing, similar to conventional pulse oximetry, with either high-end monochromatic cameras or controlled monochromatic light sources. These methods can achieve precise SpO2 measurement, but require specialized equipment. The second category uses consumer-grade RGB cameras, such as digital webcams and smartphone cameras, which are more accessible but pose challenges due to the broad wavelength range sensed by each R, G, and B channel. Previous non-contact SpO2 measurement methods using RGB cameras have primarily relied on explicit feature extraction and linear regression, but we propose a novel approach using neural networks to distill SpO2 information from all three color channels. Our approach is inspired by the optophysiological model for SpO2 measurement and utilizes deep learning to monitor SpO2 in a contactless way with regular RGB cameras. Compared to previous non-contact methods, our approach offers several advantages, including better accuracy, reduced discomfort and sanitation concerns, and broader accessibility. We demonstrate the explainability of our model by visualizing the weights for the RGB channel combinations, which also show consistency with suggested color bands used in optophysiological methods. We also analyze the impact of different skin tones and sides of the hand on SpO2 estimation.

While deep learning for SpO2 monitoring from videos is still in the early stages, our approach shows promise and has the potential to be adopted in health screening and telehealth. It offers a convenient and noninvasive way to continuously monitor SpO2, which can serve as an early warning sign of inadequate oxygenation and clinical deterioration. As such, it has the potential to improve patient outcomes and reduce healthcare costs.

%%%%%%%%%%%%%%%%%%%%%%%%%%%%%%%%%%%%%%%%%%

\section{Method}

To accurately estimate SpO2, it is important to extract color information from a person's skin, as the physiological information related to SpO2 is embedded in the reflected/reemitted light from the skin. The process involves extracting R, G, and B time series from the skin area, which are referred to as skin color signals. The skin pixels are discriminated from the background using Otsu's method, and the ROI is identified. The R, G, and B time series are generated by averaging over the values of skin pixels for each video frame. The skin color signals are then divided into 10-second segments using a sliding window with a stride of 0.2 seconds, which are used as input for neural networks. Longer segments are used to improve resilience against sensing noise. Since the segment length is much longer than the minimally required length to contain the SpO2 information for one cycle of heartbeat, a fully-connected or convolutional structure is sufficient to capture temporal dependencies without resorting to a recurrent neural network structure. pO2 prediction neural network that are understandable and interpretable, we propose the use of layer-wise relevance propagation (LRP) [8]. LRP is a popular technique for explaining the predictions of deep neural networks by attributing relevance scores to input features. The LRP algorithm assigns relevance scores to each input feature based on its contribution to the final output prediction, allowing for better understanding of which features are most important for the prediction.

In our study, we will use LRP to identify the relevant features used by the neural network for SpO2 prediction. This will help us to understand the physiological signals that are most important for accurate SpO2 estimation. Moreover, by analyzing the relevant features, we can identify potential sources of noise or bias in the model and improve the overall robustness and accuracy of the model.

In summary, model explainability and interpretability are critical in the design of SpO2 prediction systems for healthcare applications. By using LRP to identify the relevant features used by the neural network for prediction, we can improve our understanding of the underlying physiological signals and ensure that the model is trustworthy and clinically relevant.

\section{Experimental Results}

\begin{figure*}
    \centering
    \includegraphics[scale=1]{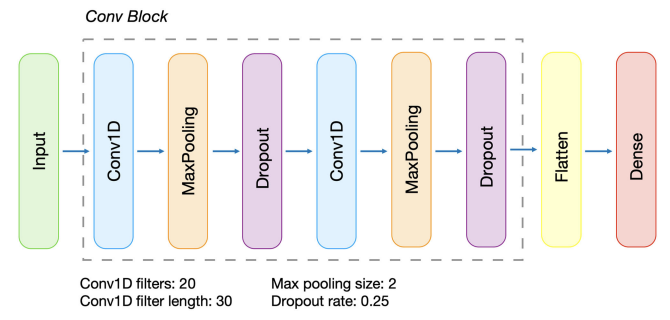}
    \caption{Parameter search using grid search for the 1D CNN.}
    \label{fig:side2side}
\end{figure*}

One of the recordings is used for training and validation of the model, while the other recording is used for testing. To ensure that there are no overlapping segments of data between the training and validation set, each recording is split into three breathing cycles, and the first two cycles are used for training, while the third cycle is used for validation.
To accurately estimate SpO2, it is important to extract color information from a person's skin, as the physiological information related to SpO2 is embedded in the reflected/reemitted light from the skin. The process involves extracting R, G, and B time series from the skin area, which are referred to as skin color signals. The skin pixels are discriminated from the background using Otsu's method, and the ROI is identified. The R, G, and B time series are generated by averaging over the values of skin pixels for each video frame. The skin color signals are then divided into 10-second segments using a sliding window with a stride of 0.2 seconds, which are used as input for neural networks.
To address the issue of the small dataset size for participant-specific experiments, the training and validation data are augmented by sampling with replacement, using the bootstrapping data reuse strategy. 
This oversampling technique also helps address the imbalance in SpO2 data values. Once the model structure and hyperparameters have been tuned using the training and validation data, multiple instances of the model are trained using the best-tuned hyperparameters. Varying the random seed used for model weights initialization and random oversampling between each instance ensures that the model is evaluated comprehensively. The model instance with the highest validation RMSE is then chosen for evaluation on the test recording.
We employed various preprocessing techniques to obtain a raw PPG signal, bandpass filtered signal, and bias signal for each color channel. To automatically extract features from these signal streams, we utilized a 1-D convolutional neural network (CNN) model, which consists of two temporal convolutional layers with max pooling and 25\% dropout. We employed a cross-validated grid search process to determine the best combination of hyperparameters, which included the number of convolutional layers, input window size, filters, and filter length. The CNN model was implemented using Python 3.5 and Tensorflow 1.2, and we trained it on a cluster of 36 nodes with an NVIDIA P100 GPU, which took approximately 4 weeks for the entire grid search process.


\begin{thebibliography}{100}

\bibitem{b1}  J. S. Zdanowicz, ‘‘Detecting money laundering and terrorist financing via
data mining,’’ Commun. ACM, vol. 47, no. 5, pp. 53–55, May 2004.
\bibitem{b1}  Johnson, E., Tang, X. and Samudrala, S., 2022. WeightMom: Learning Sparse Networks using Iterative Momentum-based pruning. arXiv preprint arXiv:2208.05970.
\bibitem{b1} H. M. Alzoubi, M. In’airat, and G. Ahmed, “Investigating the impact of total quality
management practices and Six Sigma processes to enhance the quality and reduce the
cost of quality: the case of Dubai,” Int. J. Bus. Excell., vol. 27, no. 1, pp. 94–109, 2022.
\bibitem{b1} B. Kurdi, M. Alshurideh, I. Akour, E. Tariq, A. AlHamad, and H. Alzoubi, “The effect
of social media influencers’ characteristics on consumer intention and attitude toward
Keto products purchase intention,” Int. J. Data Netw. Sci., vol. 6, no. 4, pp. 1135–1146,
2022.

\bibitem{b1} K. L. Lee, P. N. Romzi, J. R. Hanaysha, H. M. Alzoubi, and M. Alshurideh,
“Investigating the impact of benefits and challenges of IOT adoption on supply chain
performance and organizational performance: An empirical study in Malaysia,”
Uncertain Supply Chain Manag., vol. 10, no. 2, pp. 537–550, 2022.

\bibitem{b2} Subramaniam, Arvind. ”A neuromorphic approach to image processing
and machine vision,” 2017 Fourth International Conference on Image
Information Processing (ICIIP). IEEE, 2017
\bibitem{b2} Yakopcic, Chris, Raqibul Hasan, and Tarek M. Taha. "Memristor based neuromorphic circuit for ex-situ training of multi-layer neural network algorithms." In 2015 International Joint Conference on Neural Networks (IJCNN), pp. 1-7. IEEE, 2015.
\bibitem{b1} T. Sausen and A. Liegel, ‘‘AI in AML: The shift is underway,’’ NICE
Actimize, Hoboken, NJ, USA, Tech. Rep., Jan. 2020. [Online]. 
\bibitem{b1}  Estimating Illicit Financial Flows Resulting From Drug Trafficking and
Other Transnational Organized Crimes, U. N. O. Drugs and Crime,
Vienna, Austria, 2011.
\bibitem{b1}  J. Gao, Z. Zhou, J. Ai, B. Xia, and S. Coggeshall, ‘‘Predicting credit card
transaction fraud using machine learning algorithms,’’ J. Intell. Learn. Syst.
Appl., vol. 11, no. 3, pp. 33–63, 2019.
\bibitem{b1}  Y. Bengio, A. Courville, and P. Vincent, ‘‘Representation learning:
A review and new perspectives,’’ IEEE Trans. Pattern Anal. Mach. Intell.,
vol. 35, no. 8, pp. 1798–1828, Aug. 2013.
Subramaniam, A. and Sharma, A., 2019. N2NSkip: Learning Highly Sparse Networks using Neuron-to-Neuron Skip Connections.
\bibitem{b1}  J. Heaton, ‘‘An empirical analysis of feature engineering for predictive
modeling,’’ in Proc. SoutheastCon, Mar. 2016, pp. 1–6.
\bibitem{b1}  A. Coates, A. Ng, and H. Lee, ‘‘An analysis of single-layer networks in
unsupervised feature learning,’’ in Proc. 14th Int. Conf. Artif. Intell. Statist.,
JMLR Workshop Conf., 2011, pp. 215–223.
\bibitem{b1}  R. C. Watkins, K. M. Reynolds, R. Demara, M. Georgiopoulos,
A. Gonzalez, and R. Eaglin, ‘‘Tracking dirty proceeds: Exploring data
mining technologies as tools to investigate money laundering,’’ Police
Pract. Res., vol. 4, no. 2, pp. 163–178, Jun. 2003.
\bibitem{b1}  T. E. Senator, H. G. Goldberg, J. Wooton, M. A. Cottini, A. U. Khan,
C. D. Klinger, W. M. Llamas, M. P. Marrone, and R. W. Wong, ‘‘Financial
crimes enforcement network AI system (FAIS) identifying potential money
laundering from reports of large cash transactions,’’ AI Mag., vol. 16, no. 4,
p. 21, 1995.

\bibitem{b1} C. Mead. Neuromorphic electronic systems. Proceedings of the IEEE, 78(10):1629–36, 1990. 2,
10
\bibitem{b1} Henry Markram. The blue brain project. In ACM/IEEE conference on Supercomputing, SC 2006,
page 53, New York, NY, USA, 2006. IEEE, ACM. 2
\bibitem{b1} J. Schemmel, J. Fieres, and K. Meier. Wafer-scale integration of analog neural networks. In
Neural Networks, 2008. IJCNN 2008. (IEEE World Congress on Computational Intelligence).
IEEE International Joint Conference on, pages 431–438, june 2008. 2


\bibitem{b3} Rajitha, K., 2019, September. Spectral reflectance based heart rate measurement from facial video. In 2019 IEEE International Conference on Image Processing (ICIP) (pp. 3362-3366). IEEE.


\bibitem{b1} B Govoreanu, GS Kar, Y-Y Chen, V Paraschiv, S Kubicek, A Fantini, IP Radu, L Goux, S Clima,
R Degraeve, N Jossart, O Richard, T Vandeweyer, K Seo, P Hendrickx, G Pourtois, H Bender,
L Altimime, DJ Wouters, JA Kittl, and M Jurczak. 10 × 10 nm2 Hf/HfOx crossbar resistive
RAM with excellent performance, reliability and low-energy operation. International Technical
Digest on Electron Devices Meeting, pages 31–34, December 2011. 3, 6, 9

\bibitem{b3} Boppidi, P.K.R., et al., 2020. Implementation of fast ICA using memristor crossbar arrays for blind image source separations. IET Circuits, Devices \& Systems, 14(4), pp.484-489.


\bibitem{b1} Gregory S Snider and R Stanley Williams. Nano/CMOS architectures using a field-programmable
nanowire interconnect. Nanotechnology, 18(3):035204, January 2007. 4
\bibitem{b1} Tsuyoshi Hasegawa, Takeo Ohno, Kazuya Terabe, Tohru Tsuruoka, Tomonobu Nakayama,
James K Gimzewski, and Masakazu Aono. Learning Abilities Achieved by a Single Solid-State
Atomic Switch. Advanced Materials, 22(16):1831–1834, April 2010. 4
Neuromorphic nanoscale memristor synapses 19
\bibitem{b1} T. Serrano-Gotarredona, T. Masquelier, T. Prodromakis, G. Indiveri, and B. Linares-Barranco.
STDP and STDP variations with memristors for spiking neuromorphic learning systems.
Frontiers in Neuroscience, 7(2), 2013. 4, 7, 8
\bibitem{b1} C. Zamarre˜no-Ramos, L.A. Camu˜nas-Mesa, J.A. P´erez-Carrasco, T. Masquelier, T. SerranoGotarredona, and B. Linares-Barranco. On spike-timing-dependent-plasticity, memristive
devices, and building a self-learning visual cortex. Frontiers in neuroscience, 5, 2011. 4, 7



for Strategic Deterrence Studies, Air University, 2020.
\bibitem{b1} J. F. Weaver, “Regulation of artificial intelligence in the United States,” in Research
Handbook on the Law of Artificial Intelligence, Edward Elgar Publishing, 2018, pp.
155–212.
\bibitem{b1} J. Whittlestone and S. Clarke, “AI Challenges for Society and Ethics,” Oxford Handb.
AI Gov., vol. 1, no. 1, pp. 1–20, 2022, doi: 10.1093/oxfordhb/9780197579329.013.3.
\bibitem{b1} M. Shamout, R. Ben-Abdallah, M. Alshurideh, A. Kurdi, and H. B., “S. (2022). A
conceptual model for the adoption of autonomous robots in supply chain and logistics
industry,” Uncertain Supply Chain Manag., vol. 10, no. 2, pp. 577–592.
\bibitem{b1} S. M. Butt, “Management and Treatment of Type 2 Diabetes,” Int. J. Comput. Inf.
Manuf., vol. 2, no. 1, 2022.
\bibitem{b1} H. M. Alzoubi, G. Ahmed, and M. Alshurideh, “An empirical investigation into the
impact of product quality dimensions on improving the order-winners and customer
satisfaction,” Int. J. Product. Qual. Manag., vol. 36, no. 2, pp. 169–186, 2022.
\bibitem{b1} H. M. Alzoubi, H. Elrehail, J. R. Hanaysha, A. Al-Gasaymeh, and R. Al-Adaileh, “The
Role of Supply Chain Integration and Agile Practices in Improving Lead Time During
the COVID-19 Crisis,” Int. J. Serv. Sci. Manag. Eng. Technol., vol. 13, no. 1, pp. 1–11,
2022.
\bibitem{b1} M. T. Alshurideh et al., “Fuzzy assisted human resource management for supply chain
management issues,” Ann. Oper. Res., pp. 1–19, 2022.

\bibitem{b1} B. Kurdi, H. Alzoubi, I. Akour, and M. Alshurideh, “The effect of blockchain and smart
inventory system on supply chain performance: Empirical evidence from retail
industry,” Uncertain Supply Chain Manag., vol. 10, no. 4, pp. 1111–1116, 2022.

\bibitem{b3} Rajitha, K., 2019. Estimation of the Cardiac Pulse from Facial Video in Realistic Conditions.

\bibitem{b1} B. Linares-Barranco and T. Serrano-Gotarredona. Exploiting memristance in adaptive
asynchronous spiking neuromorphic nanotechnology systems. In Nanotechnology, 2009. IEEENANO 2009. 9th IEEE Conference on, pages 601–604. IEEE, 2009. 8
\bibitem{b1} S.R. Deiss, R.J. Douglas, and A.M. Whatley. A pulse-coded communications infrastructure
for neuromorphic systems. In W. Maass and C.M. Bishop, editors, Pulsed Neural Networks,
chapter 6, pages 157–78. MIT Press, 1998. 8
\bibitem{b1} E. Chicca, A.M. Whatley, P. Lichtsteiner, V. Dante, T. Delbruck, P. Del Giudice, R.J. Douglas,
and G. Indiveri. A multi-chip pulse-based neuromorphic infrastructure and its application to a
model of orientation selectivity. IEEE Transactions on Circuits and Systems I, 5(54):981–993,
2007.
\bibitem{b1} G. Eason, B. Noble, and I. N. Sneddon, ``On certain integrals of Lipschitz-Hankel type involving products of Bessel functions,'' Phil. Trans. Roy. Soc. London, vol. A247, pp. 529--551, April 1955.




\end{thebibliography}
\end{document}